\documentclass{article}

\PassOptionsToPackage{square,numbers}{natbib}
\usepackage[final]{neurips_2023}

\usepackage[utf8]{inputenc} % allow utf-8 input
\usepackage[T1]{fontenc}    % use 8-bit T1 fonts
\usepackage{hyperref}       % hyperlinks
\usepackage{url}            % simple URL typesetting
\usepackage{booktabs}       % professional-quality tables
\usepackage{amsfonts}       % blackboard math symbols
\usepackage{nicefrac}       % compact symbols for 1/2, etc.
\usepackage{microtype}      % microtypography
\usepackage{xcolor}         % colors
\usepackage{graphicx}

\title{Exploring DINO: Emergent Properties and Limitations for Synthetic Aperture Radar Imagery}

\author{%
  Joseph A. Gallego-Mejia \\
  Universidad Nacional de Colombia, Colombia\\
  \texttt{jagallegom@unal.edu.co} 
  \And
  Anna Jungbluth\\
  European Space Agency, Climate Office, UK\\
  \texttt{anna.jungbluth@esa.int} \\
  \And
  Laura Martínez-Ferrer \\
  Universitat de Val\`encia, Spain\\
  \texttt{laura.martinez-ferrer@uv.es} 
  \And
  Matt Allen \\
  University of Cambridge, UK\\
  \texttt{mja78@cam.ac.uk} 
  \And
  Francisco Dorr \\
  Independent, Argentina\\
  \texttt{fran.dorr@gmail.com} 
  \And
  Freddie Kalaitzis \\
  University of Oxford, UK\\
  \texttt{freddie.kalaitzis@cs.ox.ac.uk} 
  \And
  Raúl Ramos-Pollán \\
  Universidad de Antioquia, Colombia\\
  \texttt{raul.ramos@udea.edu.co}
}

\begin{document}

\maketitle

\begin{abstract}
  Self-supervised learning (SSL) models have recently demonstrated remarkable performance across various tasks, including image segmentation. This study delves into the emergent characteristics of the Self-Distillation with No Labels (DINO) algorithm and its application to Synthetic Aperture Radar (SAR) imagery. We pre-train a vision transformer (ViT)-based DINO model using unlabeled SAR data, and later fine-tune the model to predict high-resolution land cover maps. We rigorously evaluate the utility of attention maps generated by the ViT backbone and compare them with the model's token embedding space. We observe a small improvement in model performance with pre-training compared to training from scratch and discuss the limitations and opportunities of SSL for remote sensing and land cover segmentation. Beyond small performance increases, we show that ViT attention maps hold great intrinsic value for remote sensing, and could provide useful inputs to other algorithms. With this, our work lays the groundwork for bigger and better SSL models for Earth Observation.
\end{abstract}

\section{Introduction}

Self-supervised learning (SSL) models have gained attention in recent years for their ability to learn from unlabeled data and their knowledge transferability to unseen domains. State-of-the-art SSL models, such as Self-\textbf{Di}stillation with \textbf{No} Labels (DINO) \cite{caron2021emerging}, Contrastive Language-Image Pretraining (CLIP) \cite{radford2021learning}, and Masked Autoencoders (MAE) \cite{he2022masked}, have shown remarkable performance and generalizability in several domains. For example, DINO algorithms using vision transformer (ViT) \cite{dosovitskiy2020image} backbones have shown great performance on segmentation tasks without requiring labels \cite{caron2021emerging}.  This knowledge transferability could improve the characterization results of earth observation (EO) and remote sensing of our changing planet. In addition to optical imagery, Synthetic Aperture Radar (SAR) data, collected via active satellites, has proven immensely valuable, due to its all-weather, day-and-night monitoring capabilities. SAR imagery can be used for a variety of tasks, such as land cover segmentation, vegetation estimation, and flood detection. Leveraging the information contained in the wealth of SAR data greatly benefits from computational techniques like machine learning (ML). However, using supervised ML algorithms is challenging, due to the large amounts of labeled data required \cite{zhong2020landslide}. Here, SSL has shown great promise (e.g. \cite{zhu2021deep}).

In this work, we perform SSL on SAR imagery using ViT-based DINO models. After pre-training, we fine-tuned the model backbones to create high-resolution land cover maps. We showcase the models' ability to detect emergent features and discuss DINO's strengths and limitations when applied in this context. More specifically, we compare the performance of using ViT attention maps or token embeddings for model fine-tuning. In addition, we extensively discuss how pre-training and fine-tuning on small datasets compares to training the same architectures, or simpler supervised models, from scratch. Beyond comparing model performances, we showcase the intrinsic value and emerging segmentation of ViT attention maps when applied to large amounts of unlabelled SAR data.  With this, our work highlights the opportunities and limitations of SSL for EO and lays the groundwork for future efforts aiming to create planetary-scale foundation models for remote sensing applications.

\section{Previous Work}
Foundational models, in which a pretext task is performed, have shown enormous advantages in problems with small or sparsely labeled data sets \cite{bommasani2021opportunities}. The attention mechanism used by most foundational models has shown remarkable properties not only in large language models but in computer vision \cite{he2022masked, radford2021learning, caron2021emerging}. The potential usefulness of these foundational models could be carried over to remote sensing, where early attempts have shown promising results \cite{wang2022self, stojnic2021self, ayush2021geography}. However, most of the work on remote sensing was trained and tested with optical data, unlike the present work, where we pre-trained the models with Synthetic Aperture Radar (SAR) imagery.    

\section{Model Architecture}
In this section, we explain the pre-training task model and the top-down task model for the foundational DINO model. The pretraining task model is pre-trained using only unlabeled datasets and is subsequently fit using a top-down dataset where the dataset is assumed to be smaller compared to the pretraining dataset.

\subsection{Pre-training Task Model}

\begin{figure}[hb]
    \centering
    \includegraphics[width = \linewidth]{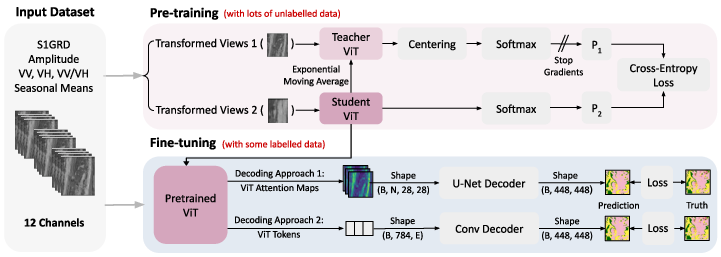}
    \caption{Schematic diagram of the ML pipeline employed in this work. We use Synthetic Aperture Radar (SAR) amplitude as input to DINO. The images undergo transformations involving multiple scaling and cropping operations before being passed to the student-teacher network. The goal of the student-teacher network is to learn the same embedding space. Centering of the teacher network is imperative to prevent model collapse. Following pre-training, the student ViT is used as the backbone for fine-tuning. Here, we compare two different approaches: we either use (1) the ViT attention maps as an input to a U-Net, or (2) the ViT token embedding space and a convolutional decoder to create high-resolution land cover maps.}
    \label{fig:dino-architecture}
\end{figure}

Figure \ref{fig:dino-architecture} shows a diagram of the ML pipeline employed in this work. The DINO model used for pre-training is structured as a teacher-student network, where the weights of the teacher are updated using an exponential moving average (ema) of the student. To preserve SAR specifics, we only perform horizontal flipping and cropping as transformations during pre-training. While many architectures can be used for the student and teacher networks, we use ViTs for their attention mechanism. 
This attention mechanism yields a token embedding space, and attention maps as a by-product. The attention maps, produced by each head of the ViT, capture distinct features and patterns in the unlabelled SAR data. 
The output of the final block of the attention mechanism is referred to as the class token. In our experiments, we omit the class token and utilize all other tokens to generate an embedding space. Both the attention maps and token embeddings can be used as input to other ML architectures.

\subsection{Downstream Task Decoders}
After pre-training, we use the student backbone to fine-tune the model for land cover segmentation. We either generate ViT attention maps as input to a simple U-Net \cite{long2015fully}, or decode the token embedding space of the ViT backbone using a convolutional decoder.  The pixel-level classification of the model is compared to the ground truth land cover maps using a cross-entropy loss. We used two different approaches to decode the embeddings created by DINO. Firstly, we use the ViT attention maps as input to a simple U-Net. Our U-Net consists of a series of down-conversions (containing 3 double convolution blocks (Conv2d, Batch Norm, Relu, Conv2d, Batch Norm, Relu) with max pooling), and a series of up-conversions (containing 4 blocks of bilinear upsampling and double convolutions), followed by a final 2D convolution to predict the 11 distinct classes of the ESAWC maps. The same U-Net is used for our baseline experiments. Secondly, we use the ViT token embedding space as encoding for predicting high-resolution land cover maps. In this case, the decoder is a series of convolutional upsampling blocks (Conv2d, Batch Norm, Relu, bilinear upsampling), to get from the token embeddings to the 448x448 pixels ESAWC predictions.

\section{Experimental Evaluation}
Three experiments were performed and are shown in this section. All experiments were tested on the ESA World Cover segmentation dataset. The first experiment uses a small pre-trained ViT-based DINO model and tests it on the ESA's dataset. The second experiment uses a larger pre-trained ViT-based DINO model and tests it on a larger portion of the ESA's dataset. The third experiment is a comparison of a pre-trained U-Net ViT-based DINO model with a supervised U-Net model.

\subsection{Dataset}

\begin{figure}[h]
    \centering
    \includegraphics[width = 0.65\linewidth]{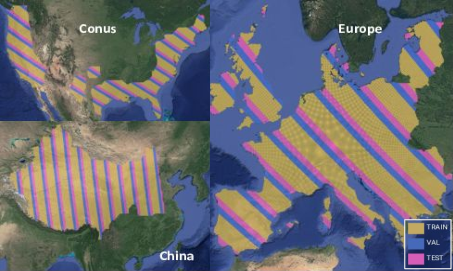}
    \caption{Data splits for CONUS, China, and Europe. There are 167K image tiles for CONUS, 285K tiles for China, and 200K tiles for Europe. We divided the data into training (60\%), validation (20\%), and test (20\%) sets based on geographic bands to minimize data leakage across contiguous tiles.}
    \label{fig:data-splits}
\end{figure}

We prepared ML-ready datasets of SAR imagery and high-resolution land cover maps for three areas of interest (AOIs): Europe, China, and the Continental United States (CONUS). SSL pre-training was either performed just in Europe or all three AOIs. All models were fine-tuned in Europe, using small labeled datasets.  The data was partitioned into tiles (448$\times$448 pixels), using \texttt{geetiles}\footnote{\url{https://github.com/rramosp/geetiles}} and \texttt{sartiles}\footnote{\url{https://github.com/rramosp/sartiles}}, and all tiles were divided into training (60\%), validation (20\%), and test (20\%) sets based on geographic bands (see Figure \ref{fig:data-splits} in the SI). To keep the data volume manageable, we only used data from 2020.
We processed SAR amplitude from the \href{https://developers.google.com/earth-engine/datasets/catalog/COPERNICUS_S1_GRD}{Sentinel-1 Level 1 Ground Range Detected} (S1GRD) dataset. SAR amplitude measures the intensity of vertically (V) or horizontally (H) polarized pulses and back-scatter. As input to our model, we calculated and stacked seasonal averages of VV, VH, and the logarithmic difference (VV-VH), totaling 12 channels. To prepare our labeled dataset for model fine-tuning, we used the \href{https://esa-worldcover.org/en}{ESA's World Cover} (ESAWC) dataset. ESAWC provides global land cover maps at 10 m resolution and contains 11 distinct land cover classes.

\begin{figure}[h]
    \centering
    \includegraphics[width = 0.7\linewidth]{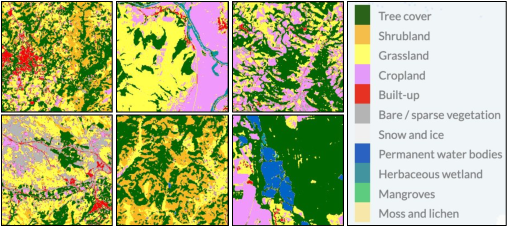}
    \caption{Example \href{https://esa-worldcover.org/en}{ESA World Cover} maps. Each map contains pixel-level classification across 11 distinct classes. The maps are derived from Sentinel 1 and Sentinel 2 imagery. Each tile measures 448x448 pixels, with a spatial resolution of 10 meters.}
    \label{fig:example-esawc-maps}
\end{figure}

\subsection{Experiment 1: Benefits \& Limitations of Small Pre-trained ViT-based DINO model}

To investigate the impact of SSL for downstream land cover segmentation, we pre-trained two models using the DINO architecture; a model with ViT Tiny backbones (3 attention heads, 6.1 million parameters) was trained on Europe alone, while a model with ViT Base backbones (12 attention heads, 88 million parameters) was trained on data from Europe, CONUS, and China. All ViTs encoded information in patches of size 16x16 pixels, leading to attention maps of 28x28 pixels.

Throughout our fine-tuning experiments, we optimized the binary cross-entropy between the model outputs and ground truth ESAWC maps and reported the mean intersection over the union (MIOU) as the performance metric. Table \ref{table:attention-maps-vs-tokens-europe} (top) presents the results of land cover segmentation when using attention maps or token embeddings as the encoding. The model was either trained from scratch or using the ViT backbone pre-trained in Europe. In order to investigate model performances when labeled datasets are limited, we fine-tuned all models on 0.1\%, 1\%, or 10\% of available data in Europe.

\paragraph{Results and discussions}
Using attention maps, we observe a 4\% (3.5\%) improvement in MIOU with the pre-trained ViT compared to training from scratch on 0.1\% (1\%) of data. However, with 10\% of available data, we no longer see a performance improvement with pre-training. In contrast, when using the token embedding space as encoding, we observe a more substantial improvement of 9\% (9.5\%) in MIOU with pre-training compared to training from scratch on 0.1\% (1\%) of data. Similar to the attention map approach, this difference disappears when using 10\% of the fine-tuning data. 

Notably, models with a frozen backbone, where the ViT component is not updated during fine-tuning, consistently yield inferior results. Unsurprisingly, performances increase when more data is used for fine-tuning. The token embeddings generally lead to better performances than when attention maps are used as encoding. However, the maximum 9.5\% improvement in MIOU remains relatively modest, begging the question whether the added computational cost of pre-training is worth this small improvement.

\subsection{Experiment 2: Large foundational pre-trained ViT-based DINO model}
We next explore whether larger ViTs pre-trained on data from diverse geographic regions improve the separability between pre-training and training from scratch. Table \ref{table:attention-maps-vs-tokens-europe-conus-china} (bottom) presents the results obtained when pre-training a larger ViT-based DINO model on S1GRD images from Europe, CONUS, and China.  

\paragraph{Results and discussions}
Notably, with just 0.1\% of fine-tuning data, training from scratch yields better results than using the pre-trained backbone for both encoding approaches. The pre-trained ViT with attention map encoding demonstrates a 4\% (2\%) improvement over training from scratch when fine-tuning on 0.1\% (10\%) of data. In contrast, the pre-trained model with token embeddings exhibits a more substantial 14\% (10\%) improvement over training from scratch with 1\% (10\%) of data. 

We hypothesize that using a larger ViT may require a larger fine-tuning dataset to fully leverage the information learned during pre-training. In line with the results presented for pre-training on Europe, freezing the pre-trained backbone of the multi-region model consistently leads to inferior results. 

\subsection{Experiment 3: Pretrain Foundational model against Supervised model}
In order to fully evaluate the performance of our models, we compare our approach to a supervised U-Net baseline, where instead of using the attention maps or token embeddings as input, we directly translate the 12-channel S1GRD SAR imagery to ESAWC maps. 

\paragraph{Results and discussions}
Table \ref{table:unet-baselines} shows that adding attention maps as an additional input further improves the MIOU, leading to the overall best results presented here.  Interestingly, for small dataset sizes (<10\%) a simple U-Net baseline leads to much better results. With 10\% of fine-tuning data, equally good results are achieved for the U-Net baseline and our larger ViT using token embeddings. We therefore hypothesize that the benefits of pre-training become more apparent when larger pre-training and fine-tuning datasets are used.

\begin{table}[h!]
\centering
\caption{Comparison of using attention maps or token embeddings for predicting ESAWC maps, when pre-training on only Europe (top table; ViT Tiny) or Europe, CONUS, and China (bottom table; ViT base). The mean intersection over the union (MIOU) was calculated across the entire test set. All models were fine-tuned on 0.1\%, 1\%, or 10\% of training data, for 100 - 200 epochs.}

\begin{tabular}{c||c|c|c||c|c|c}
    \textbf{Europe} & \multicolumn{3}{|c||}{Attention Maps} & \multicolumn{3}{|c}{Tokens} \\
    \hline
    & 0.1\% & 1\% & 10\% & 0.1\% & 1\% & 10\% \\
    \hline
    From scratch & 0.278 & 0.348 & \textbf{0.398} & 0.248 & 0.377 & 0.430 \\
    \hline
    With pre-training (backbone frozen) & 0.160 & 0.209 & 0.249 & 0.245 & 0.317 & 0.403 \\
    \hline
    With pre-training (backbone not frozen) & \textbf{0.290} & \textbf{0.360} & 0.396 & \textbf{0.275} & \textbf{0.395} & \textbf{0.431} \\
\end{tabular}\label{table:attention-maps-vs-tokens-europe}

\bigskip

\begin{tabular}{c||c|c|c||c|c|c}
    \textbf{Europe, CONUS, China} & \multicolumn{3}{|c||}{Attention Maps} & \multicolumn{3}{|c}{Tokens} \\
    \hline
    & 0.1\% & 1\% & 10\% & 0.1\% & 1\% & 10\% \\
    \hline
    From scratch & \textbf{0.310}& 0.387 & 0.449 & \textbf{0.293} & 0.354 & {\color[HTML]{FE0000} 0.476}\\
    With pre-training (backbone frozen) & 0.205 & 0.324 & 0.369 & 0.254 & 0.338 & 0.436\\
    With pre-training (backbone not frozen) & 0.267 & \textbf{0.403} & \textbf{0.460} & 0.286& \textbf{0.403}& \textbf{0.477}
\end{tabular}\label{table:attention-maps-vs-tokens-europe-conus-china}

\end{table}

\begin{table}[h]
\centering
\label{table:unet-baselines}
\caption{Comparison of a supervised U-Net baseline, to a U-Net with attention maps as an additional input. The reported metric is the intersection of the union (MIOU). The U-net baseline was trained from scratch on 12 channels of S1GRD (VV, VH, VV/VH for each season) to predict ESAWC maps. This is compared to a U-Net trained on 15 channels, i.e, 12 channels of S1GRD, plus 3 attention maps from a vision transformer. The vision transformer was trained on 100\% of training data in Europe.}
\begin{tabular}{c||c|c|c||c|c|c}
    & \multicolumn{3}{|c||}{U-Net} & \multicolumn{3}{|c}{U-Net} \\
    & \multicolumn{3}{|c||}{(only S1GRD)} & \multicolumn{3}{|c}{(S1GRD + Attention Maps)} \\
    \hline
    & 0.1\% & 1\% & 10\% & 0.1\% & 1\% & 10\% \\
    \hline
    From scratch & 0.376 & 0.412 & 0.475 & \textbf{0.380} & 0.426 & \textbf{0.480} \\
    With pre-training (backbone frozen) & - & - & - & 0.371 & 0.430 & 0.474 \\
    With pre-training (backbone not frozen) & - & - & - & 0.378 & \textbf{0.432} & 0.477 \\
\end{tabular}
\end{table}

\begin{figure}
    \centering
    \includegraphics[width=0.98\linewidth]{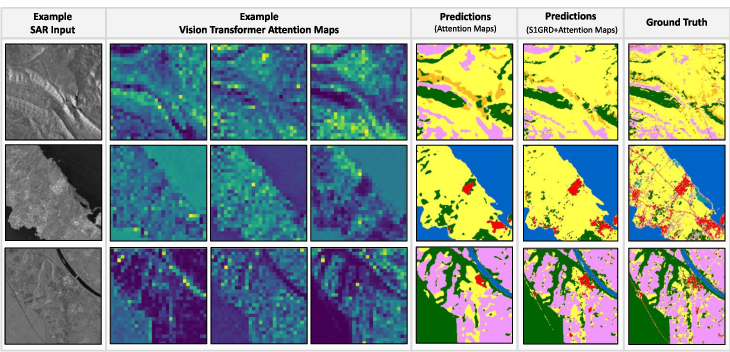}
    \caption{Example SAR input (column 1), ViT attention maps (columns 2-4), land cover predictions (columns 5-6), and ground truth ESAWC maps. The predictions were either generated using only attention maps, or S1GRD + attention maps as input to a U-Net. Both models were pre-trained and fine-tuned on 10\% of data in Europe, without freezing the backbone. For all land cover maps, the different colors correspond to different classes, as indicated in the SI.}
    \label{fig:input, predictions, ground truth}
\end{figure}

\section{Conclusions}
In this manuscript, we have drawn several noteworthy conclusions on the use of SSL for remote sensing and land cover segmentation. Firstly, we found that pre-training ViT-based DINO models leads to small improvements in downstream task performance compared to training from scratch, especially when using the ViT tokens embedding space as encoding. Using only ViT attention maps as input for fine-tuning shows a limited ability to produce details in high-resolution land cover maps. Here, the choice of ViT patch size emerges as a crucial factor, with smaller patch sizes leading to higher-resolution attention maps and potentially better results for tasks that require great spatial detail. Nonetheless, attention maps could provide great value as inputs for other ML algorithms. Notably, while a supervised U-Net baseline generally outperformed our ViT-based encoding-decoding architectures, the integration of attention maps as an additional input yields promising results, indicating the potential for synergistic effects. While further experimentation into SSL for EO is needed, particularly for planetary-scale foundational models, we see attention maps as a useful tool for achieving fully explainable and generalizable models, especially when applying learned knowledge to diverse downstream tasks with limited labeled data.

\section{Acknowledgements}

%Hidden for double blind review

This work has been enabled by Frontier Development Lab Europe (\url{https://fdleurope.org}) a public/private partnership between the European Space Agency (ESA), Trillium Technologies, the University of Oxford and leaders in commercial AI supported by Google Cloud and Nvidia, developing open science for all Humankind.  L.M-F. was supported by the European Research Council (ERC) Synergy Grant “Understanding and Modelling the Earth System with Machine Learning (USMILE)” under the Horizon 2020 research and innovation programme (Grant agreement No. 855187). M. J. A. was supported by the UKRI Centre for Doctoral Training in Application of Artificial Intelligence to the study of Environmental Risks [EP/S022961/1], and additionally by Trinity Hall, Cambridge. We are also indebted to Nicolas Longépé, Carlos López-Martínez, Fabio A. González Osorio, Samuel Bancroft, Emma Hatton, Alison Lowndes, Alistair Francis, Ioanna Bouri and the rest of reviewers during 2023 FDL-Europe sprint. 

\clearpage
\bibliography{references}

\clearpage
\section{Supplementary Information}
\begin{table}[h!]
    \centering
        \caption{Overview of the training parameters used for DINO pre-training using ViT tiny or ViT base.}
        \label{tab:dino-parameters}
        \begin{tabular}{c|c|c}
        \hline
         Parameter &  ViT Tiny & ViT Base\\
         \hline
         Center Momentum & 0.99 & 0.99 \\
         Embedding Dimension & 192 &  768 \\
         Number of Heads & 3 & 12 \\
         Learning Rate & 0.000001 & 0.000001 \\
         Student Temperature & 0.03 & 0.03 \\
         Teacher Temperature & 0.001 & 0.001 \\
         Warm-up Teacher Temperature & 0.01 & 0.01 \\
         Warm-up Teacher Epochs & 5 & 5 \\
         Model Parameters & 6.1 M & 88.8 M\\
         \hline
    \end{tabular}
\end{table}

\begin{figure}
    \centering
    \includegraphics[width=0.7\linewidth]{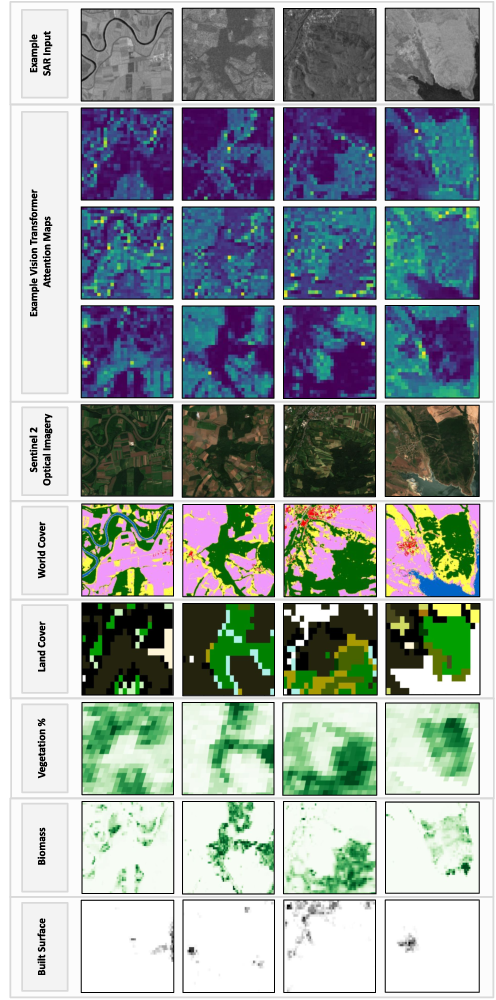}
    \caption{Comparison of ViT attention maps and various satellite imagery and derived data products. The SAR input (row 1) was used to pre-train the ViT. Looking at 3 example attention maps (rows 2-4) we clearly see emergent segmentation. Each attention head focuses on different features and patterns in the data. When comparing these features to other types of data, such as Sentinel 2 optical imagery (row 5), \href{https://esa-worldcover.org/en}{ESA World Cover} maps (row 6), \href{https://climate.esa.int/en/projects/land-cover/about/}{ESA CCI Medium Resolution Land Cover} maps (row 7), estimations of \href{https://lpdaac.usgs.gov/products/mod44bv006/}{vegetation percentage} (row 8), \href{https://climate.esa.int/en/projects/biomass/}{biomass} (row 9) or \href{https://ghsl.jrc.ec.europa.eu/download.php?ds=bu}{built surface} (row 10), we clearly see that many of the features detected by the ViT translate to prominent features in the various downstream datasets.}
    \label{fig:attention-maps-downstream-tasks}
\end{figure}
\subsection*{Intrinsic Value of Attention Maps.}
To visually inspect the results, Figure \ref{fig:input, predictions, ground truth} shows example SAR inputs, ViT attention maps, model predictions, and ground truth ESAWC maps. Column 4 shows the predictions using just the attention maps as input, while column 5 shows the results for a model trained with S1GRD + attention maps. 

Firstly, we observe that the ViT attention maps clearly capture features and patterns in the SAR data. Secondly, we see that predictions using S1GRD + attention maps show greater detail than if only attention maps are used as input to the U-Net. All attention maps span 28x28 pixels, which is likely too small to predict high-resolution land cover maps with great detail. While the ViT patch size could in principle be reduced to obtain higher-resolution attention maps, this would greatly increase the required memory during training. 

However, even at low resolutions, the ViT attention maps clearly contain valuable information derived from the SAR input data. Figure \ref{fig:attention-maps-downstream-tasks} shows how well features and patterns detected by the ViT match satellite imagery (e.g. Sentinel 2) and derived data products for a variety of downstream tasks (e.g. vegetation, biomass, or built area estimation). Consequently, we believe that attention maps are useful inputs to other ML algorithms, and will prove valuable tools for model explanability and generalizability in the context of SSL or EO.

\end{document}